\documentclass[12pt]{article}
\usepackage[latin1]{inputenc}
\usepackage[T1]{fontenc}
\usepackage[english]{babel}
\usepackage{ae}
\usepackage{amsmath,amssymb,amstext}
\usepackage{graphicx,cite,a4wide,array}
\usepackage{booktabs}
\usepackage{subfigure}

\begin{document}

\thispagestyle{empty}



\begin{center}

\LARGE\bf Empirical review of\\ standard benchmark functions\\
  using evolutionary global optimization\\[2cm]
   \Large\it Johannes M.~Dieterich and Bernd Hartke$^\ast$\\ [1cm]
   \large\rm $\dagger$ Institut f\"ur Physikalische Chemie,\\
             Christian-Albrechts-Universit\"at,\\
             Olshausenstra\ss e 40,\\
             D--24098 Kiel, Germany\\[5mm]
             $\ast$ corresponding author, hartke@pctc.uni-kiel.de
\end{center}




\vspace{1cm}

\section*{Abstract}

We have employed a recent implementation of genetic algorithms to study a
range of standard benchmark functions for global optimization. It turns out
that some of them are not very useful as challenging test functions, since
they neither allow for a discrimination between different variants of
genetic operators nor exhibit a dimensionality scaling resembling that of
real-world problems, for example that of global structure optimization of
atomic and molecular clusters. The latter properties seem to be simulated
better by two other types of benchmark functions. One type is designed to be
deceptive, exemplified here by Lunacek's function. The other type offers
additional advantages of markedly increased complexity and of broad
tunability in search space characteristics. For the latter type, we use an
implementation based on randomly distributed Gaussians. We advocate the use of
the latter types of test functions for algorithm development and benchmarking.

\vspace{2cm}

{\bf keywords:} global optimization, genetic algorithms, benchmark functions,
dimensionality scaling, crossover operators

\newpage

\setcounter{page}{1}

\section{Introduction}
\label{section:intro}

Global optimization has a lot of real-world applications, both of discrete and
non-discrete nature. Among them are chemical applications such as structure
optimization of molecules and clusters, engineering problems such as component
design, logistics problems like scheduling and routing, and many others.
Despite the typical practical finding that a general global optimization
algorithm usually is much less efficient than specific versions tuned to the
problem at hand, it is still of interest to gauge the baseline performance of
a global optimization scheme using benchmark problems. Even in the most recent
examples of such tests \cite{ozer,thangaraja,hirsch,zhao,suganthan,bbob}
(selected at random from
the recent literature), it is customary to employ certain standard
benchmark functions, with the implicit (but untested) assumption that the
difficulty of these benchmark functions roughly matches that of real-world
applications. Some of these benchmark functions even are advertised as
particularly challenging. 

We have developed evolutionary algorithm (EA) based global optimization
strategies in the challenging, real-life area of atomic and molecular cluster
structure optimization \cite{my1stGA,phenix,tip4p_n,bb,ogolem1,ljclusters}.
When we apply our algorithms to those traditional, abstract
benchmark functions, however, neither of those two claims
(challenge, and similarity to real-world applications) stands up. In fact,
similar suspicions have been voiced earlier. For example, already in 1996
Whitley \textit{et al.} \cite{whitley} argued that many of
the standard benchmark functions should be relatively easily solvable due to
inherent characteristics like symmetry and separability. Some functions even
appeared
to get easier as the dimensionality of the function increases. Nevertheless,
as the citations in the previous paragraph indicate, the same set of
traditional benchmark functions continues to be used indiscriminately to the
present day, by the
majority of researchers in various fields. Therefore, with the present
article, we address the need to re-emphasize those earlier findings from a
practical point of view, add in
other test functions, and extend the testing to higher dimensionality. In
addition,
we stress the conclusions that these
traditional benchmark problems appear to be too simple to allow for meaningful
comparisons between algorithms or implementation details, and that they do not
allow conclusions about the performance of global optimization algorithms in
real-life situations. We show that the latter is achieved better when using
different kinds of benchmark functions.

In these contexts, theoretical considerations often focus on classifying a
problem
as $N$ or $N\!P$ \cite{wille,greenwood}, or on evaluation of marginally
different
parameter
representations \cite{salomon,whitley} or hybrid combinations of known test
functions \cite{whitley}.
Additionally, translations and rotations as well as randomized noise are
typically added to the normal benchmark functions\cite{bbob}. Albeit being
a potentially
viable approach to complicate the benchmark, it constrains the comparability
of different benchmark results by introducing additional parameters and thereby
incompatibilities. As this is of central importance when discussion
independently
developed, novel algorithms, this discussion will not contain benchmark
functions
modified in this manner.
Quite independent of such problem classifications and
algorithm characteristics, however, in most real-world applications
scaling with problem dimension (i.e., number of parameters to be optimized)
plays a pivotal role. Chemical structure optimization of clusters is an
obvious example: Of central practical importance are phenomena like cluster
aggregation and fragmentation, or the dependence of properties on cluster
size, while isolating a single cluster size is a formidable experimental
challenge. Therefore, one does not study a single cluster size but tries to
systematically study a range of clusters
\cite{phenix,takeuchi_lj,tip4p_n,takeuchi_water,roy_book}, only limited by the
maximum computing capacity one has. It is obvious that smallest decreases in
scaling (e.g.\ from $\mathcal{O}(N^{3.5})$ to $\mathcal{O}(N^3)$) may allow
for significantly larger cluster sizes to be studied.

The problem dimensionality scaling of the number of global optimization steps
needed is of course linked to features of the global optimization
algorithm. For evolutionary algorithms, this includes crossover and mutation
operators, possible local optimization steps and problem-specifically tuned
additional operators. We would like to present here latest results of some
standard benchmark functions in the context of our recently developed
framework for the evolutionary global optimization of chemical problems,
{\sc ogolem} \cite{ogolem1}. By screening the needed amount of global minimizing
steps for solving up to 500 (in one case 10000) dimensional benchmark
functions, we obtain the scaling of different crossover operators with the
dimensionality of these functions.  Additionally, we compare runs with and
without local optimization steps in some cases to investigate the effect of
gradient based
minimization on the scaling. Last but not least, we compare the performance on
these standard benchmark functions with that on different kinds of benchmark
function that 
apparently present more serious challenges, coming closer to real-world
problems in some respects.

The present work
contributes to defining
a new baseline and standard for
benchmarking
global optimization algorithms. By demonstrating how a modern implementation of
genetic algorithms scales in solving the reviewed benchmark functions, we want
to
encourage other developers of global optimization techniques to report not only
results for a particular dimensionality of a defined benchmark function but
focus
on the scaling behaviour and compare their results to our empirical baseline.

\section{Methods and Techniques}
\label{section:methods}

All calculations mentioned in this paper were carried out using our {\sc ogolem}
framework \cite{ogolem1} written in Java with SMP parallelism enabled. Since
differing concurrency conditions can obviously have an impact on the
benchmarking results, all calculations were carried out with 24 concurrent
threads.

{\sc ogolem} is using a genetic algorithm (GA), loosely based on the standard GA
proposed in Ref.~\cite{goldbergbook} but differing in the treatment of the
genetic population. Instead of a classical generation based global
optimization scheme, a pool algorithm \cite{bb} is used. This has the
advantage of both eliminating serial bottlenecks and reducing the number of
tunables since, e.g., elitism is build-in and no rate needs to be
explicitly specified. 

Tunables remaining with this approach are mentioned in table
\ref{tab_tunables} with values kept constant in the benchmark runs.

\begin{table}[ht]
\begin{center}
\begin{tabular}{clc}
\toprule
\textit{Tunable} & \textit{Representation}
& \textit{value} \\
\textit{(pool approach)} & \textit{(generation approach)} & \\
\midrule
pool size & generation size & 1000 \\
global optimization steps & generation size times& till
minimum is reached \\
& number of generations & \\
fitness diversity & threshhold which individuals
& $1\cdot10^{-8}$\\
& are considered to be same & \\
\bottomrule
\end{tabular}
\caption{Tunables in the pool algorithm and their value during the benchmark.}
\label{tab_tunables}
\end{center}
\end{table}

The genetic operators are based upon a real number representation of the
parameters. Within the crossover operator, the cutting is genotype-based. 
Different crossover operators used below differ only in the numbers and
positions of cuts through the genome. The positions are defined by randomized
number(s) either being linearly distributed or Gaussian distributed\footnote{We
are using in both cases the standard PRNG provided by the Java Virtual Machine
(JVM) and defined by the Java standard.}, with
the maximum of the Gaussian being in the middle of the genome and with the
resulting Gaussian-distributed random numbers multiplied with 0.3 to make the
distribution sharper. In Tab.~\ref{algotab} the used algorithms are summarized
and explained.

\begin{table}[ht]
\begin{center}
\begin{tabular}{llll}
\toprule
\textit{Algorithm} & \textit{Crossing} & \textit{Number of Crossings} &
\textit{Crossing Point}\\
\midrule
Holland & no & 0 & N/A\\
Germany & yes & 1 & Gaussian\\
Portugal:1 & yes & 1 & linear\\
Portugal:3 & yes & 3 & linear\\
Portugal:5 & yes & 5 & linear\\
Portugal:7 & yes & 7 & linear\\
\bottomrule
\end{tabular}
\end{center}
\caption{Definition of the used algorithms.}
\label{algotab}
\end{table}

Mutation and mating are the same for all algorithms and tests. The mutation is a
standard one-point genotype mutation with a probability of 5\%. The actual gene
to be mutated is chosen with a linearly distributed random number and replaced
with a random number in between allowed borders specific to every
function/application/parameter.

Mating is accomplished by choosing two parents from the genetic pool. The
mother is chosen purely randomly, whilst
the father is chosen based on a fitness criterion. All structures in the pool
are ranked by their fitness, a Gaussian distributed random number shaped with
the factor 0.1 is chosen with its absolute value mapped to the rank in the pool.

If a local optimization step is carried out, it is a standard L-BFGS, as described
in Ref. \cite{lbfgspaper,nocedal} and implemented in RISO\cite{riso}, with very tight convergence criteria
(e.g.
$1\cdot10^{-8}$ in fitness). The needed gradients are analytical in all cases.

Once the crossing, mutation and (if applicable) local optimization steps have
been carried out on both children, only the fitter one will be returned to the
pool. This fitter child will actually be added to the pool if it has a lower
function value than the individual with the highest function value in the pool
and does not violate the fitness diversity criterion. The fitness diversity
is a measure to avoid premature convergence of the pool. Additionally, it
promotes exploration by maintaining a minimum level of structural diversity,
indirectly controlled via fitness values. For example, by assuming that two
individuals with the same fitness (within a threshold) are same, it eliminates
duplicates. Since the pool size stays constant, the worst individual will be
dropped automatically, keeping the pool dynamically updated.

For the benchmarking, we do not measure timings since they are of course
dependent upon convergence criteria of the local optimization and potentially
even on the exact algorithm used for the local optimization (e.g.\ L-BFGS vs.
BFGS vs.\
CG).
Therefore, our benchmarking procedure measures the number of global
optimization steps where a step is defined to consist of mating, crossing,
mutation and (if applicable) local optimization. The second difficulty is to
define when the global optimum is found. We are using the function value as a
criterion, trying to minimize the amount of bias one introduces by any measure.
It should be explicitly noted here that within the benchmarking of a given
test function, this value stays constant in all dimensionalities which should in
principle increase the difficulty with higher dimensionalities, again minimizing
the amount of (positive) bias introduced. For all local tests with local
optimization
enabled, five independent runs have been carried out and a standard deviation is
given.

\section{Standard benchmark functions}
\label{section:benchmark}

Any approach towards global optimization should be validated with a set of
published benchmark functions and/or problems. In the area of benchmark
functions a broad range of published test functions exists, designed to stress
different parts of a global optimization algorithm. Among the most popular ones
are Schwefel's, Rastrigin's, Ackley's, Schaffer's F7 and Schaffer's F6
functions. They have the
strength of an analytical expression with a known global minimum and, in the
case of all but the last function, they are extendable to arbitrary
dimensionality
allowing for scaling investigations. Contrary to assumptions made frequently,
however, most of these benchmark functions do not allow to discriminate between
algorithmic variations in the global optimization, nor do they give a true
impression of the difficulty to be expected in real-life applications, as we
will demonstrate in the following subsections.

\subsection{Ackley's Function}

\begin{figure}[ht]
\centering
\subfigure[Full search space]{
\includegraphics[width=7cm]{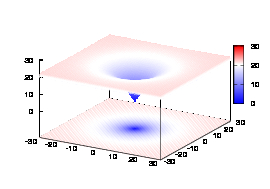}}
\subfigure[Fine structure]{
\includegraphics[width=7cm]{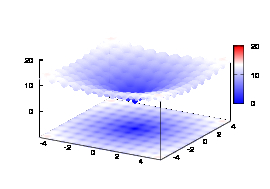}}
\caption{2D plot of Ackleys function.}
\label{plot_ackley2d}
\end{figure}

Ackley's function has been first published in Ref.~\cite{ackley1} and has been
extended to arbitrary dimensionality in Ref.~\cite{ackley2} It is of
the form
\begin{equation}
  \mathop{f(\vec{x})}=-20\cdot
      \exp\left[-0.2\sqrt{\frac{1}{n}\cdot\sum\limits_{i=1}^{n}x_{i}^{2}}\right]
      - \exp\left[\frac{1}{n}\cdot\sum\limits_{i=1}^{n}\cos(2\pi
      x_{i})\right] + 20 + e^1
\end{equation}
with the global minimum at $x_i=0.0$. We considered this to be a relatively
trivial function due to its shape consisting of a single funnel (see fig.
\ref{plot_ackley2d}). Nevertheless, this function type potentially has
relevance for real-world applications since e.g.\ the free energy hypersurface
of
proteins is considered to be of similar, yet less symmetric, shape.

The initial randomized points were drawn from the interval
\begin{equation}
 -32.768\leq x_i \leq 32.768
\end{equation}
for all $x_i$, which is to our knowledge the normal benchmark procedure for this
function.

\begin{figure}[ht]
\centering
\subfigure[]{
\includegraphics[width=7cm]{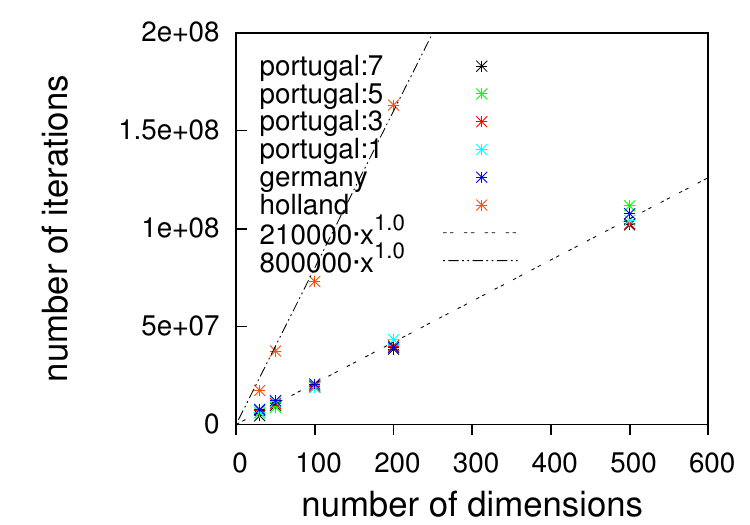}}
\subfigure[]{
\includegraphics[width=7cm]{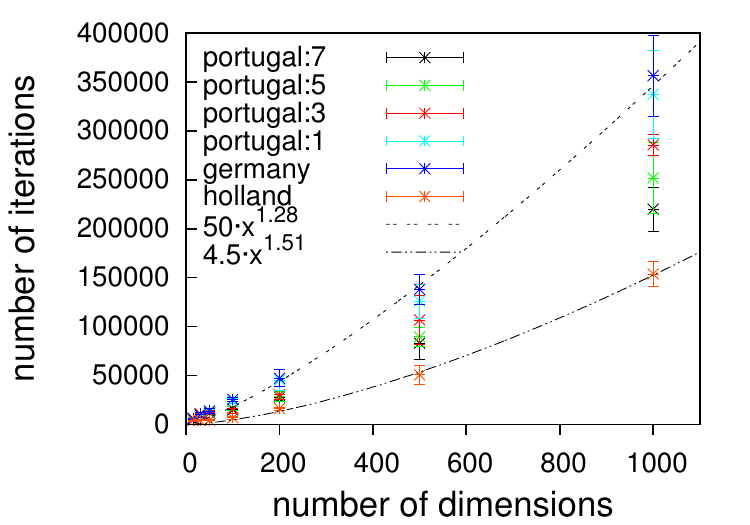}}
\caption{Scaling results for Ackley's function. a) without, b) with local
optimization.}
\label{ackley_results}
\end{figure}

As can be seen from Fig.~\ref{ackley_results}, without local optimization
steps the choice of genotype operator
makes almost no difference; all cases exhibit excellent 
linear scaling. The only deviating case is the mutation-only
algorithm, \emph{Holland}, which has a higher prefactor in
comparison to the other algorithms but still exhibits the same (linear)
scaling. With local optimization enabled, the
results do vary more. Results up to 500 dimensions do not allow for a concise
statement on the superiority of a specific crossover operator. We therefore
extended the benchmarking range up to one thousand dimensions,
hoping for a clearer picture. It should be noted here that
on a standard contemporary 24-core compute node, these calculations took
3.5 minutes in average (openJDK7 on an openSUSE 11.4), 
demonstrating the good performance of our framework.

Even with the extended benchmarking range, no concise picture could be
obtained. This most likely means that none of the used algorithms is clearly
superior to the others, in agreement with the results gained from the
runs without local optimization. The only difference remaining is that the
prefactor of the mutation-only algorithm is reduced to equality with the
crossover algorithms. In general, the cases with enabled local optimization do
increase the scaling slightly from 1.0 up to 1.28 and 1.51 for Germany and
Holland,
respectively. Nevertheless, even this increased scaling is still excellent and
the
usage of local optimization steps proves to significantly lower the time to
solution.

Still, these results obviously allow for the conclusion that Ackley's benchmark
function should be considered to be of trivial difficulty since linear scaling
is
achievable already without local optimization. Any state-of-the-art global
optimization
algorithm should be capable of solving it with a scaling close to linear.

\begin{figure}[ht]
\centering
\subfigure[Exact gradient]{
\includegraphics[width=7cm]{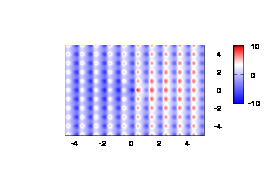}}
\subfigure[Simplified gradient]{
\includegraphics[width=7cm]{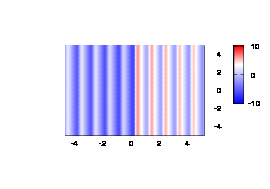}}
\caption{Contour plot of gradient element $x_1$ for a 2D-Ackley's function.}
\label{plot_ackleygrad}
\end{figure}

Also, we want to demonstrate an interesting aspect of Ackley's benchmark
function when manipulating the analytical gradient expression in a distinct
manner. In general the modification of  gradients in order to simplify the
problem is not unheard of in applications of global optimization techniques to
chemical problems (see e.g. Ref.~\cite{guntramberndfit}) and we therefore
consider it to be of interest.
The analytical gradient for a gradient element $i$ is defined as
\begin{equation}
g(i)=4\cdot x_i\dfrac{\exp{\left(-0.2\sqrt{a}\right)}}{n\sqrt{a}}
+\dfrac{2\pi\sin{(2\pi x_i)}\exp{(b)}}{n}
\end{equation}
with
\begin{equation}
a=\dfrac{\sum_{i=1}^n x_i^2}{n}
\end{equation}
and
\begin{equation}
b=\dfrac{\sum_{i=1}^n \cos{(2\pi x_i)}}{n}.
\end{equation}
The simple modification that
\begin{equation}
a=\dfrac{x_i^2}{n}
\end{equation}
and
\begin{equation}
b=\dfrac{\cos{(2\pi x_i)}}{n}
\end{equation}
does not only drastically simplify the gradient expression, but also
further helps to simplify the global optimization problem. As can be seen in
Fig.~\ref{plot_ackleygrad},
this simplification effectively decouples the dimensions, thereby smoothing the
search space.

With this simplification in place, Ackley's function can be solved by simply
locally optimizing
a couple of randomized individuals (the first step of our global optimization
algorithms) up to 10000 dimensions with constant effort.
 
\begin{figure}[ht]
\centering
\includegraphics[width=7cm]{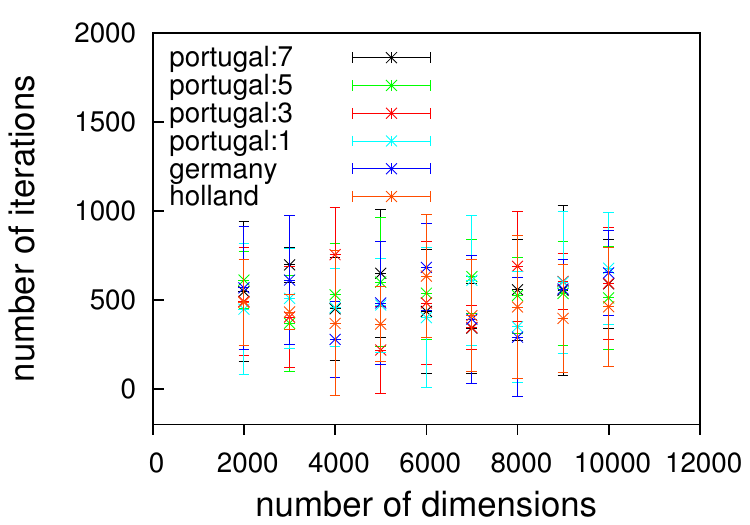}
\caption{Scaling behaviour of Ackley's function with a modified gradient
expression.}
\label{ackley_optresults}
\end{figure}

\subsection{Rastrigin's Function}

Rastrigin's function\cite{rastrigin1,rastrigin2} does have fewer minima within
the defined search space of
\begin{equation}
 -5.12\leq x_i \leq 5.12
\end{equation}
but its overall shape is flatter than Ackley's function which should complicate
the general convergence towards the global optimum at $x_i=0.0$.

We defined Rastrigin's function with an additional harmonic potential outside
the search space to force the solution to stay within those boundaries when
using unrestricted local optimization steps.

\begin{equation}
  f(x_0 ... x_n) =  10\cdot n + \sum_{i=0}^{n}
  \begin{cases}
  x_i > 5.12\vee x_i < -5.12: & 10\cdot x_i^2
\\[2ex]
-5.12\leq x_i \leq 5.12: & x_i^2 - 10\cdot\cos(2\pi x_i)
  \end{cases}
  \end{equation}

\begin{figure}[ht]
\centering
\includegraphics[scale=1.0]{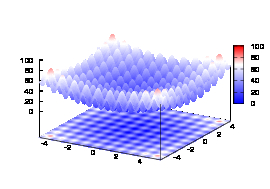}
\label{plot_rastrigin2d}
\caption{2D plot of Rastrigin's function}
\end{figure}

\begin{figure}[ht]
\subfigure[]{
\includegraphics[width=5cm]{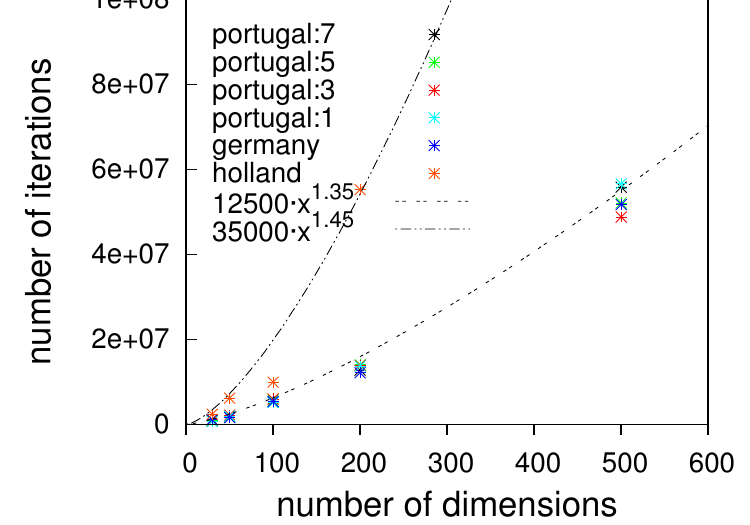}}
\subfigure[]{
\includegraphics[width=5cm]{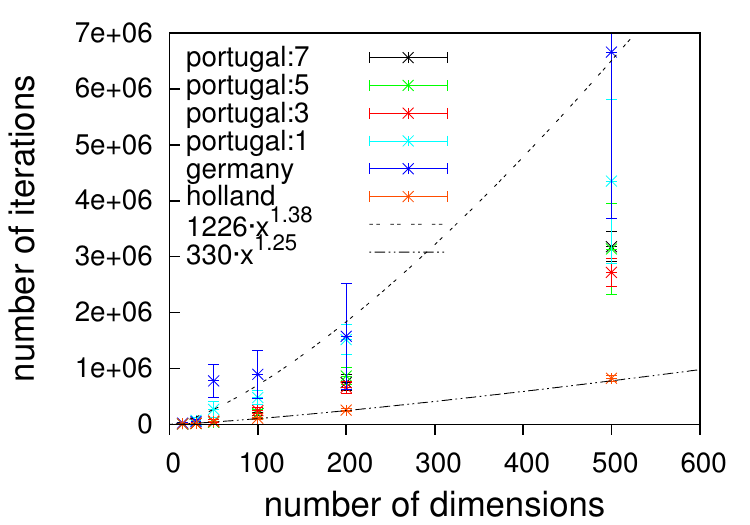}}
\subfigure[]{
\includegraphics[width=5cm]{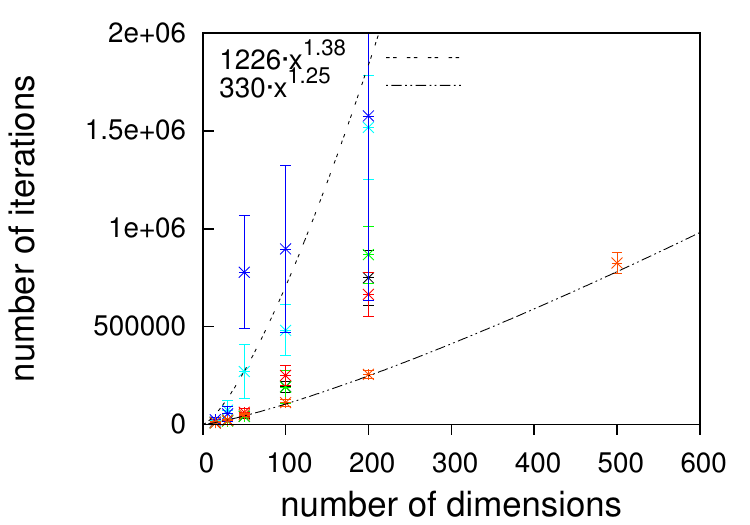}}
\caption{Scaling results for Rastrigin's function. a) without, b) with local
optimization and c) zoom with local optimization.}
\label{rastrigins_results}
\end{figure}

As can be seen from Fig.~\ref{rastrigins_results}, the scaling is excellent
with all tested crossing operators; \emph{Holland} again being the easily
rationalizable exception, in the case without local optimization. In the case
with local optimization, a contrasting picture can be seen.

The scaling once more does not deviate much between the different algorithms
with the prefactor making the major difference. Interestingly, by far the best
prefactor and scaling is obtained with \emph{Holland}. Probably, the rather
non-intrusive behaviour of a mutation-only operator fits this problem best
since it provides a better short-range exploration than any of the crossing
algorithms. We assume the rather high number of global optimization steps
(also in comparison to the non-locopt case) to be due to a repeated finding of
the same, non-optimal minima. Inclusion of \emph{taboo-search} 
features \cite{tabu_engels} into the
algorithm might be of help for real-world problems of such a type, not
reducing the 
number of global optimization steps but the amount of time spent in
local optimizations rediscovering already known minima.

Just as Ackley's function discussed earlier, it is possible to conclude that
Rastrigin's function should be solvable with
almost linear scaling using contemporary algorithms.

\subsection{Schwefel's Function}

In comparision to Rastrigin's function, Schwefel's function\cite{schwefel}
adds the difficulty
of being less symmetric and having the global minimum at the edge of the search
space
\begin{equation}
-500.0 \leq x_i \leq 500.0
\end{equation}
at position $x_i=420.9687$. Additionally, there is no overall, guiding slope
towards the
global minimum like in Ackley's, or less extreme, in Rastrigin's function.

Again, we added an harmonic potential around the search space
\begin{equation}
            f(x_0...x_n)=418.9829\cdot n + \sum_{i=0}^{n}
\begin{cases}
x_i > 500\vee x_i < -500: & +0.02\cdot x_i^2
\\[2ex]
-500\leq x_i \leq 500: & -x_i\cdot\sin\left(\sqrt{|x_i|}\right)
\end{cases}
  \end{equation}
since otherwise our unrestricted local optimization finds lower-lying minima
outside the 
principal borders.

\begin{figure}[ht]
\centering
\includegraphics[scale=1.0]{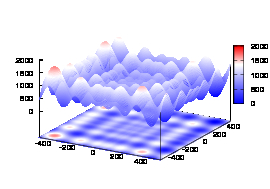}
\caption{2D plot of Schwefels function}
\label{plot_schwefel2d}
\end{figure}

\begin{figure}[ht]
\subfigure[]{
\includegraphics[width=7cm]{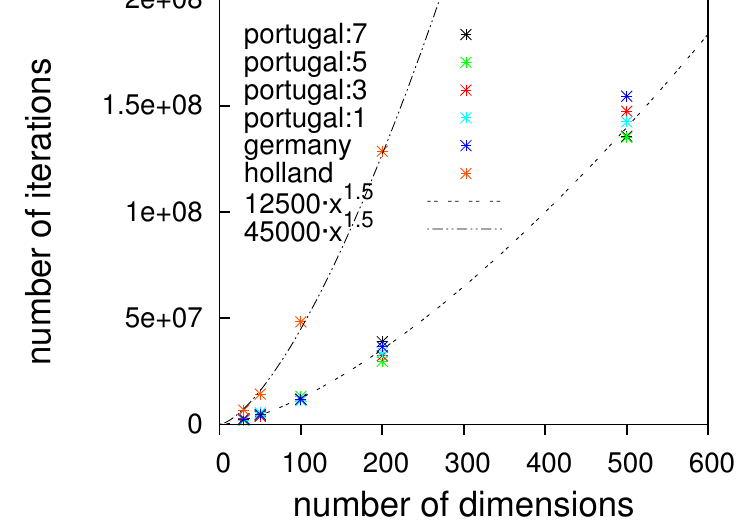}}
\subfigure[]{
\includegraphics[width=7cm]{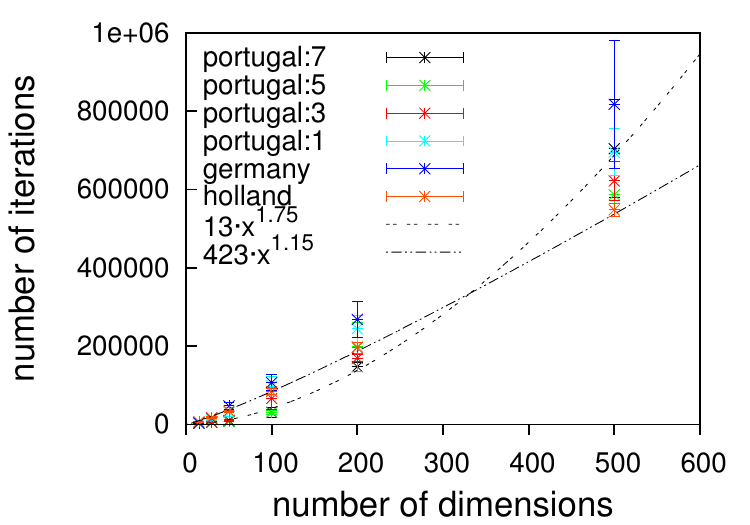}}
\caption{Scaling results for Schwefel's function. a) without, b) with local
optimization.}
\label{schwefels_results}
\end{figure}

As can be seen from Fig.~\ref{schwefels_results}, sub-quadratic scaling can
be achieved with and without local optimization. Once more, without local
optimization the non-crossing algorithm has a higher prefactor but the same
scaling
as the others, which is equalized when turning on local optimizations. In this
particular case, the usage of local optimization steps has the potential to
slightly
affect the scaling from 1.15 to 1.75. 

Again, we must come to the conclusion that a sub-quadratic scaling is far better
than what we would expect to obtain for real-world problems, restricting the
usage of Schwefel's
function as a test-case for algorithms designed to solve the latter.

\subsection{Schaffer's F7 function}
Schaffer's F7 function is defined as
\begin{equation}
f(x_0...x_n)=\left[\dfrac{1}{n-1}\sqrt{s_i}\cdot\left(\sin{(50.0s_i^{\frac{1}{5}
})}+1\right)\right]^2
\end{equation}
with
\begin{equation}
s_i=\sqrt{x_i^2+x_{i+1}^2}
\end{equation}
in $n$-dimensional space within the boundaries
\begin{equation}
-100.0 \leq x_i \leq 100.0
\end{equation}

\begin{figure}[htb]
\centering
\subfigure[Full search space]{
\includegraphics[width=7cm]{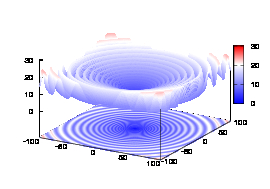}}
\subfigure[Fine structure]{
\includegraphics[width=7cm]{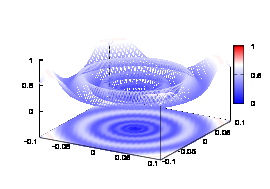}}
\caption{Plot of Schaffer's F7 function}
\label{plot_schafferf7}
\end{figure}

\begin{figure}[ht]
\subfigure[all algorithms]{
\includegraphics[width=7cm]{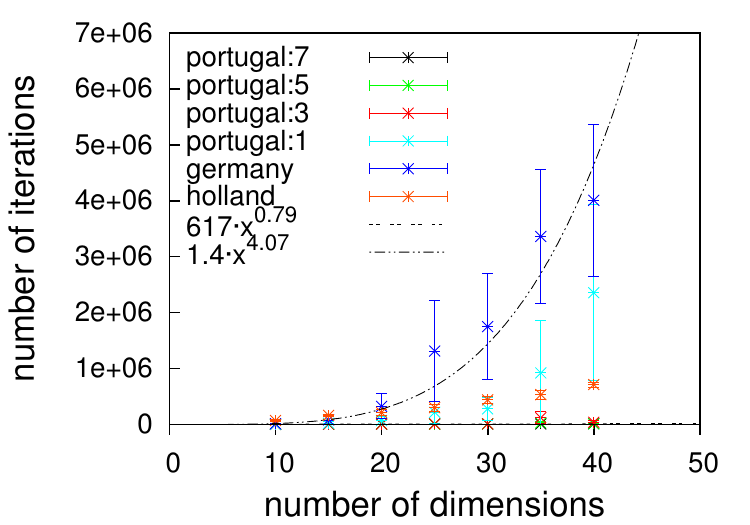}}
\subfigure[zoom]{
\includegraphics[width=7cm]{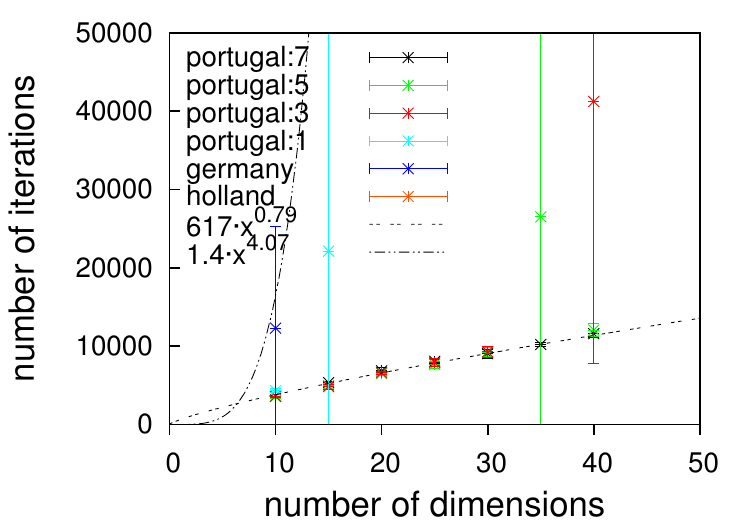}}
\caption{Scaling results for Schaffer's F7 function with local
optimization.}
\label{schafferf7_results}
\end{figure}

As can be seen from Fig.~\ref{plot_schafferf7}, concentric barriers need to be
overcome
in order to reach the global minimum $x_i=0.0$. Judging from the results
depicted in Fig.~\ref{schafferf7_results},
this function is capable of discriminating between different algorithmic
implementations. For the Gaussian-based single point crossover (\emph{germany}) the
scaling is of quartic nature, whilst all the other algorithms are scaling
linearly with the problem size. The sublinear scaling is an artifact of the high
ratio of crossover cuts (up to seven) to problem size (only 40D) and it can be expected to
increase to linear for higher dimensionalities. Interestingly, the performance
of the mutation-only algorithm again is on par with the algorithms employing crossover
operators. We see such a bias with all the benchmark functions so far, and
this does not conform to our experience with real-life problems. Therefore,
conclusions on the importance of mutation for global optimization may be too positive
when based on an analysis using only these functions.

Nevertheless, we consider Schaffer's F7 function the most interesting of the benchmark
functions studied so far. It should be explicitly noted though that a forty-dimensional
problem size will definitely be too small when exclusively analyzing algorithms
employing multiple crossover cuts.

\subsection{Schaffer's F6 Function}

For completeness, we would like to present some non-scaling results using
Schaffer's F6 function as a benchmark:

\begin{equation}
  f(x,y)=0.5+\dfrac{\sin^2(\sqrt{x^2+y^2})-0.5}{\left[
    1+0.001\cdot(x^2+y^2)\right]^2 }
  \end{equation}

\begin{figure}[htb]
\centering
\subfigure[Full search space]{
\includegraphics[width=7cm]{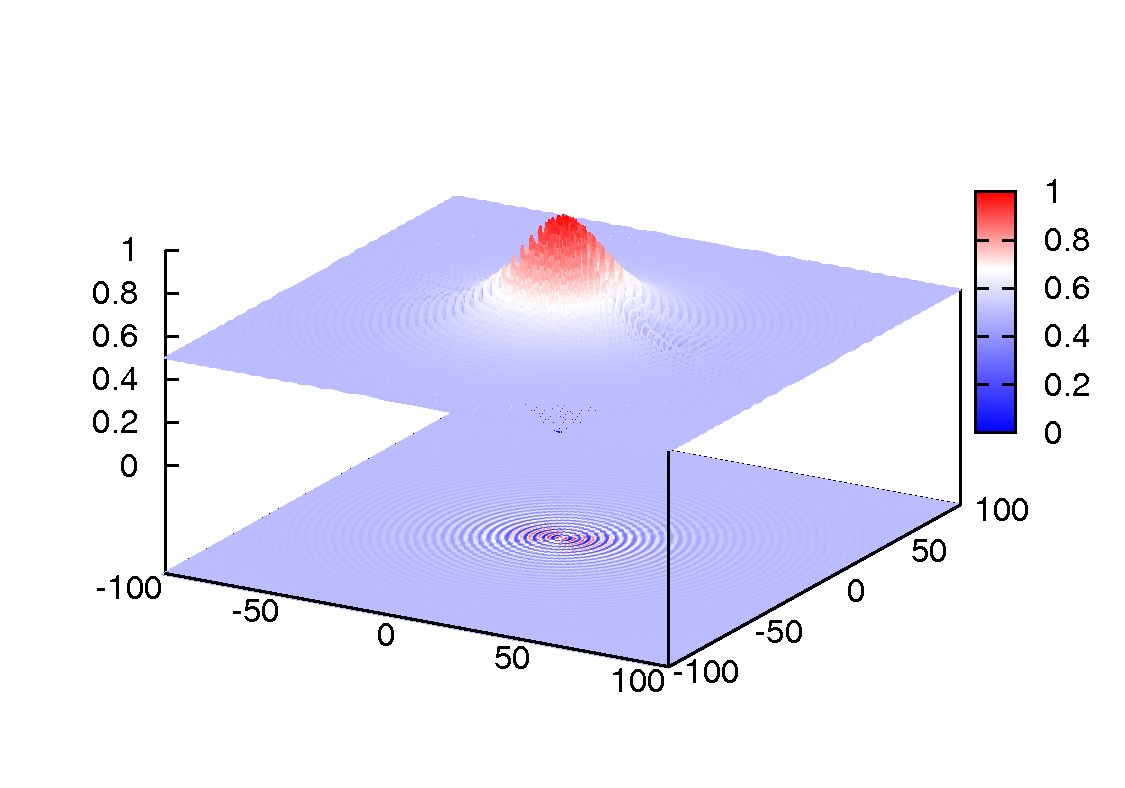}}
\subfigure[Fine structure]{
\includegraphics[width=7cm]{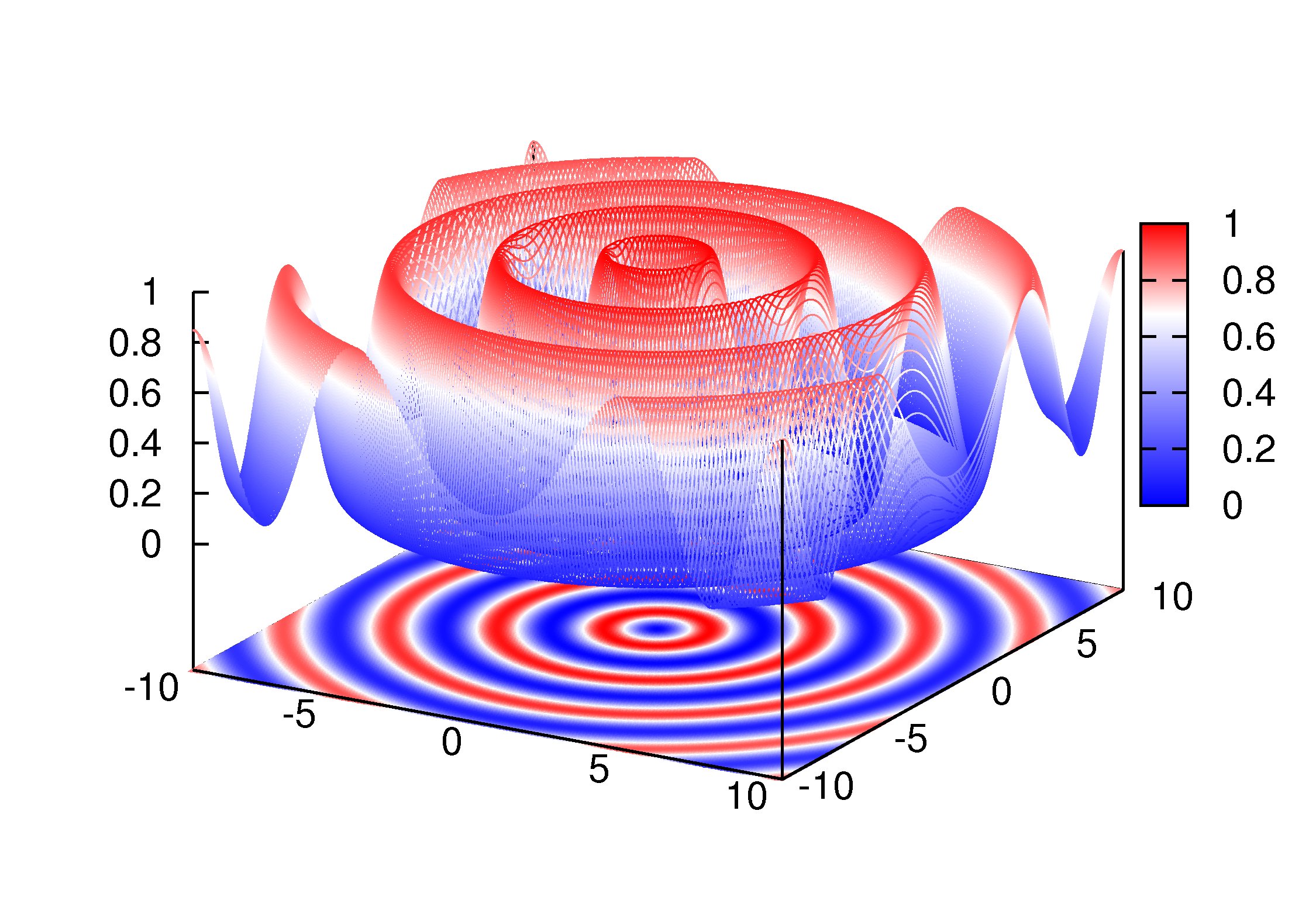}}
\caption{Plot of Schaffer's F6 function}
\label{plot_schaffer}
\end{figure}

As can be seen in Fig.~\ref{plot_schaffer}, the difficulty in this function is
that the size of the potential maxima that need to be overcome to get to a
minimum increases the closer one gets to the global minimum.

In Tab.~\ref{tab_schaffer}, results can be found which were obtained with
\emph{Holland} and with the one-point crossover operators (obviously, with a
real-number encoded genotype approach, not more cuts can take place for a
two-dimensional function). The results are outcomes of three successive runs
which is sufficient to obtain a general picture of the trend.

\begin{table}[ht]
\begin{center}
\begin{tabular}{lrr}
\toprule
\emph{Algorithm} & \emph{w/ locopt} & \emph{w/o locopt}\\
\midrule
Holland     & 990$\pm$325   & 180284$\pm$134848 \\
Germany     & 1901$\pm$1282 & 1609516$\pm$1905293 \\
Portugal:1  & 620$\pm$473   & 832919$\pm$436006 \\
\bottomrule
\end{tabular}
\end{center}
\caption{Different results for Schaffers F6 function with one-point and
zero-point crossover operators.}
\label{tab_schaffer}
\end{table}

The difference between the \emph{Portugal} and \emph{Germany} approach in this
very special case is that \emph{Germany} in contrast to \emph{Portugal} can
also yield a crossover point before the first number, effectively reducing it to
a partial non-crossing approach.

Interestingly, we do see converse tendencies between the case with and without
local optimization. We see a clear preference of the \emph{Germany} crossover
operator over \emph{Portugal} without local optimization. Taking the results of
the non-crossing operator into account, it seems clear that without local
optimization too much crossing is harmful in terms of convergence to the global
minimum. With local optimization enabled, these differences disappear,
sometimes even allowing the global minimum to be found in the initial (and
therefore never crossed) pool of solutions.

Although Schaffer's function shows an impressive difficulty for a
two-dimensional function, it should still be easily solvable with and without
local
optimization.

\subsection{Scatter of the benchmarking results}

Obviously any stochastic approach is difficult to benchmark in a reliable
manner. Therefore, we would like to discuss the scatter of the benchmarking
results. We will try to approximate possible deviations for every crossover
operator used with and without local optimization for a two hundred dimensional
Ackley's function. For this, we present in Tab.~\ref{tab_deviations} results
from ten runs per crossover operator used.

\begin{table}[ht]
\begin{center}
\begin{tabular}{llrrrrr}
\toprule
\textit{Algorithm} & \emph{Local opt.}&\textit{Maximum} & \textit{(\% dev.)} &
\textit{Minimum} & \textit{(\% dev.)} &
\textit{Average}\\
\midrule
Holland & w/ &   8991 &  18.5 &  6113 & 19.5 & 7590 \\
	& w/o&  10323905  & 13.0  & 7805543  & 14.5  & 9132168 \\
Germany & w/ &  11289  & 57.1  & 4758  & 33.8 & 7186\\
	& w/o&  4312566  & 28.1  & 2704586  & 19.6	 & 3365305 \\
Portugal:1 & w/& 20464 & 126.8  &  5191 & 42.5 & 9023  \\
	& w/o &  4748780 & 36.7  & 2095005  & 39.7 & 3473084 \\
Portugal:3 & w/& 7172 & 27.6  & 4704  & 16.3 & 5621 \\
	& w/o &  4640199 & 45.5  &  2328333 & 27.0 & 3189287 \\
Portugal:5 & w/& 6268 & 20.2  &  4873 & 6.6 & 5215\\
	& w/o &  4745643 & 39.4  & 2160887  & 36.5 & 3403532 \\
Portugal:7 & w/& 6510 & 15.3  & 4858  & 14.0  & 5646 \\
	& w/o & 4221855  & 29.9  &  2541700 & 21.8 & 3249622\\
\bottomrule
\end{tabular}
\end{center}
\caption{Number of global optimization iterations from ten
successive runs on the 200D Ackley function.}
\label{tab_deviations}
\end{table}

Of course, the results do not and cannot take all or the maximum possible
deviations into account since in principle there should be a probability
distribution from a single iteration up to infinity which can only be captured
adequately by an infinite amount of successive runs. Nevertheless, ten
successive runs can be considered to give a rough idea of the location of the
maximum.

The impression gained in the previous sections, namely that the exact nature of
the genotype operator does not seem to make a difference when using local
optimization, holds true also with enhanced statistics for this case. Similarly,
the differences seen in the runs without local optimizations between the
crossing operators and the non-crossing operator, \emph{Holland}, remain
also when averaging over more runs.

This allows for the conclusion that the results presented in the previous
sections are giving a reasonably accurate picture, despite of course
suffering from the inherent uncertainty in all stochastic methods which cannot
be circumvented.

Upon closer examination of the data in Table \ref{tab_deviations}, some
seemingly systematic trends can be observed, calling for speculative
explanations. 
A general tendency observed is the reduced spread when local optimization is
turned off, probably because a higher diversity can be maintained
providing a better and more reliable convergence. Another tendency is the
reduced spread when more --- or no --- crossover points are used. In the case of
more crossover points this can be explained with more crossover points causing
bigger changes in each step; this improves search space coverage, which in
turn makes the runs more reproducible. For the reduced scatter of the
non-crossing operator, the
explanation is obviously the opposite, since this operator minimizes changes
to the genome, allowing for a better close-range exploration.

Despite of these interesting observations, we refrain from further analysis
since this would lead us outside of the scope of the present article.

\section{Gaussian benchmark class}
\label{section:gaussians}

To our experience from the global optimization of
chemical systems, real-world problems are considerably more challenging than
the benchmark functions described above. For example,
in the case of the relatively trivial Lennard-Jones (LJ) clusters, the best
scaling
we could reach is cubic \cite{phenix}. Therefore, we feel a need for benchmarks
with
a difficulty more closely resembling real-world problems.

Defining new benchmark functions is of course not trivial since they
should fulfill certain criteria.
\begin{enumerate}
  \item Not trivial to solve.
  \item Easy to extend to higher dimensions causing higher difficulty.
  \item Possibility to define an analytical gradient for gradient based methods.
  \item Of multimodal nature with a single and known global minimum.
\end{enumerate}
To have better control over these criteria when generating benchmark
functions, a few ``search landscape generators'' have been proposed in recent
years \cite{gallagher,locatelli,gaviano}. The simplest and most flexible of
these is the one based on randomly distributed Gaussians
\cite{gallagher}. For convenience, we have used our own implementation of this
concept, abbreviated \emph{GRUNGE} (GRUNGE: Randomized, UNcorrelated Gaussian
Extrema), defined and discussed in the following. We would like to emphasize
already at this point that our intention in using \emph{GRUNGE} is not to
re-iterate known results from Ref.~\cite{gallagher} and similar work, but to
directly contrast the OGOLEM behavior displayed in section
\ref{section:benchmark} with its different behavior in the \emph{GRUNGE}
benchmark. This shows strikingly that the rather uniform results in section
\ref{section:benchmark} are not a feature of OGOLEM but rather a defect of
that benchmark function class.

We define a function as a set of randomized Gaussians
\begin{equation}
  f(x_0 ... x_M)=\sum^{N}_{i=0}\xi_i\cdot\exp\left[-\zeta_i
\cdot\sum^{M}_{j=0}\left(x_j-\kappa_j\right)^2\right]
\end{equation}
with the random numbers $\xi_i$, $\zeta_i$ and $\kappa_i$ being the weight,
width and position of the $i$-th Gaussian in $M$-dimensional space. As can be
easily seen, this class of benchmarking problems provides (besides the search
space size) two degrees of freedom. One is the number $N$ of randomized
Gaussians in the search space and $M$ being the dimensionality of the
Gaussians.\footnote{As a side note, we write \emph{GRUNGE($M$,$N$)},
e.g. for 2000 gaussians in a ten dimensional space \emph{GRUNGE[10,2000]}.} More
subtly, there is also a connection between these two
characteristics and the Gaussian widths within the maximal coordinate interval
(i.e., the Gaussian density). With proper choices of these numbers, one can
smoothly tune such a benchmark function between the two extremes of a
``mountain range'' (many overlapping Gaussians) and a ``golf course''
(isolated Gaussians with large flat patches in-between).

Functions in this class of benchmarks are not easy to solve, easily extendable
in
dimensionality and --- through the use of Gaussians --- the definition of an
analytical gradient is trivial. The only problem remaining is to pre-determine
the position and depth of the global minimum. Here, other benchmark function
generators like, e.g., the polynomial ones proposed by Gaviano \textit{et al.}
\cite{gaviano} and Locatelli \textit{et al.} \cite{locatelli}, 
allow for more control, but at the price of more uniform
overall features of the generated test functions, which is exactly what we
want to avoid. We also do not want to enforce a known global minimum by
introducing a single,
dominating Gaussian with excessively large weight by hand. Thus, the only
remaining
possibility is to define a fine grid over the search space and to run local
optimizations starting at every grid point, to obtain a complete
enumeration of all minima within the search space. Due to the simple
functional form and to the availability of an analytical gradient, this is a
realistic proposition for moderately-dimensioned examples (10D) containing a
sufficiently great number of sufficiently wide Gaussians.

In contrast to the traditional benchmarks examined in the previous sections,
the \emph{GRUNGE} function is not deliberately designed to be
deceptive, in any number of dimensions. Instead, due to the heavy use of random
numbers in its definition, it does not contain any correlations whatsoever. To
our experience, this feature makes \emph{GRUNGE} benchmarks much harder than
any of the traditional benchmarks. We cannot offer formal proofs at this
stage, but our distinct impression from many years of global optimization
experience is that realistic problems tend to fall in-between these two
extremes, being harder than traditional benchmarks but less difficult (less
uncorrelated) than \emph{GRUNGE}.

Obviously, a full exploration of the randomize Gaussians set of benchmark
functions requires an exclusive and extensive study, which has already been
started by others \cite{gallagher}. As already mentioned above, our sole
intention here is to provide a contrast to the OGOLEM behavior noted in
section \ref{section:benchmark}. To this end, we present results
based on solving a ten-dimensional \emph{GRUNGE} benchmark with
2000 gaussians (\emph{GRUNGE[10,2000]}) within a search space of
\begin{equation}
0.0 \leq x_i \leq 10.0
\end{equation}
with local optimization enabled.

\begin{table}[ht]
\begin{center}
\begin{tabular}{lrrrr}
\toprule
\textit{Algorithm} & \textit{Run 1} & \textit{Run 2} & \textit{Run 3} &
\textit{Average}\\
\midrule
Holland    & 5725 & 7676 & 6228 & 6543 \\
Germany    & 2390 &  153 & 5390 & 2644 \\
Portugal:1 & 3145 & 7637 & 4077 & 4953 \\
Portugal:3 & 1879 & 2575 & 1339 & 1931 \\
Portugal:5 & 2441 & 1647 & 4888 & 2992 \\
Portugal:7 & 5533 &  369 & 6322 & 4074 \\
\bottomrule
\end{tabular}
\end{center}
\caption{GRUNGE[10,2000] benchmarking results with local optimization steps.}
\label{tab:grungewithlo}
\end{table}

As can be seen from the results in Tab.~\ref{tab:grungewithlo}, the average
of three independent runs of all algorithms yields results within the same
order of magnitude. When comparing the numbers in Tab.~\ref{tab:grungewithlo}
with the results given above for the conventional benchmark functions, e.g.,
with those in Tab.~\ref{tab_deviations}, one should remember the differences
in dimensionality: Here we are dealing with a 10-dimensional problem with 2000
minima, whereas in Tab.~\ref{tab_deviations} we reported the performance on
the 200-dimensional Ackley function with the number of minima being several
orders of magnitude higher. This gives an indication of what we experience as a
big difference in difficulty.

It should be noted, however, that in two cases the
global optimum could be found within the locally optimized initial parameter
sets. While this demonstrates once more that a randomly distributed initial
parameter set can
have an extraordinary fitness, it also indicates that
higher dimensional \emph{GRUNGE} benchmarks are necessary to better emulate
real-world problems.

We also did some tests without local optimization, showing that the
\emph{GRUNGE} benchmark with our randomly generated Gaussians is extremely
difficult to solve without local optimization, requiring almost 19 million
global optimization steps with \emph{Portugal:3}. We suspect that 
this level of difficulty is
related to inherent features of the \emph{GRUNGE} benchmark (e.g., to the
completely missing correlation between the locations and depths of the minima)
but also to features of the specific GRUNGE[10,2000] incarnation used here
(i.e., this particular Gaussian distribution and density), but decide to leave
this sidetrack at this point.

\section{Lunacek's function}
\label{sec:lunacek}

Lunacek's function\cite{lunacek}, also known as the  bi- or double-Rastrigin
function, is a
hybrid function consisting of a Rastrigin and a double-sphere part and is
designed
to model the double-funnel character of some difficult LJ cases, in particular
LJ$_{38}$.

\begin{equation}
\label{eq:Luna}
f(x_0...x_n)=\min\left(\left\{\sum_{i}^N(x_i-\mu_1)^2\right\},\left\{d\cdot
N+s\cdot\sum_i^N(x_i-\mu_2)^2\right\}\right)+10\sum_i^N(1-\cos2\pi(x_i-\mu_1))
\end{equation}
\begin{equation}
\mu_2=-\sqrt{\dfrac{\mu_1^2-d}{s}}
\end{equation}

This indicates that there is an interest in developing benchmark functions
of higher difficulty, and indeed the developed function provides an interesting
level
of difficulty, as we show below. Nevertheless, we would like to dispute the
notion that it resembles certain real-world problems and the source of their
difficulty. Specifically,
Lunacek \emph{et al.} claim that the global
optimization of homogeneous LJ clusters is one of the most important
applications of global optimization in the field of computational chemistry.
Furthermore, they claim that the most difficult instances of the LJ problem
possess a double-funnel landscape. 
The former claim is a rather biased view and promotes
the importance of homogenous LJ clusters from a mere benchmark system
to a hot-spot of current reasearch. 
As the broad literature on global cluster structure optimization documents
(cf.\ reviews\cite{Angew,StructBond,ogolem1,WIREs}
and references cited therein), 
current challenges in this field rather are directed towards
additional complications in real-life applications, e.g., how to taylor search
steps to dual-level challenges of intra- and intermolecular conformational
search in clusters of flexible molecules, or how to reconcile the
vast number of necessary function evaluations with their excessive cost at the
ab-initio quantum chemistry level. In terms of search difficulty, the
homogeneous LJ case is now recognized as rather easy for most cluster sizes,
interspersed with a few more challenging problem realizations at certain
sizes, with LJ$_{38}$ being the smallest and hence the simplest of them. This
connects to the second claim by Lunacek \emph{et al.}, namely that the
difficulty of LJ$_{38}$ arises from the double-funnel shape of its search
landscape, which is captured by their test function design. It is indeed
tempting to conclude from the disconnectivity graph analysis by Doye, Miller
and Wales \cite{WalesDoyeDisconnect} that there are two funnels, a narrow one
containing the fcc global minimum, separated by a high barrier from the broad
but less deep one containing all the icosahedral minima. Even if this were
true (to our knowledge, such a neat separation of the two structural types in
search space has not been shown), it would
give rise to only two funnels in a 108-dimensional search space, which is not
necessarily an overwhelming challenge and also not
quite the same as what eq.\ \ref{eq:Luna} offers.

Lunacek's function is specifically designed to poison global optimization
strategies working with bigger population sizes. This is achieved through the
double sphere contribution which constructs in every dimension a fake minimum,
e.g., when using the settings $s=0.7$, $d=1.0$, with the optimal
minimum located at $x_i=2.5$. As Lunacek \emph{et al.} have proven in their
initial publication, the function is very efficient in doing this. This is an
observation that we can support from some tests on 30-dimensional cases. 

\begin{table}[ht]
\begin{center}
\begin{tabular}{llrr}
\toprule
\textit{Dimensionality} & \emph{Static grid}& \textit{Steps to solution} &
\textit{MNIC}\\ 
\midrule
 2 & w/o & 833      & N/A \\
   & w/  & 1063     & 250 \\
 5 & w/o & 2186     & N/A \\
   & w/  & 4184     & 100 \\
10 & w/o & 392006   & N/A \\
   & w/  & 13922    & 100 \\
15 & w/o & 459317   & N/A \\
   & w/  & 604954   & 50  \\
20 & w/o & not found& N/A \\
    & w/ & 1153811  & 50  \\
30 & w/o & not found& N/A \\
    & w/ & 2826707  & 20  \\
\bottomrule
\end{tabular}
\end{center}
\caption{Exemplary benchmarking results of the Lunacek function. All results 
obtained with local optimization steps and the germany algorithm. Not found
corresponds to more than 10 million unsuccessful global optimization steps. MNIC
is the maximum number of individuals allowed per grid cell.}
\label{tab_lunacekwithlo}
\end{table}

Clearly, this is a markedly different behavior than that observed above for
Ackley's, Rastrigin's, Schwefel's or Schaffer's functions, coming closer to
what we experienced in tough application cases. Therefore, it is not
surprising that additional measures developed there are also of some help
here. One possibility is to adopt a niching strategy, similar to what was
applied to reduce the solution expense for the tough cases of homogenous LJ
clusters to that of the simpler ones \cite{phenix}. In essence, this ensures a
minimum amount of diversity in the population, preventing premature collapse
into a non-globally optimal solution.

The most trivial implementation of niching is to employ a static grid over
search space and to allow only a certain maximal population per grid cell (MNIC,
maximum number of individuals per grid cell). Already this trivial change allows
the previously unsolvable function to be solved in 30 dimensions, as can be seen
from Tab.~\ref{tab_lunacekwithlo}. Solving higher dimensionalities (e.g. 100D)
with this approach suffers again from dimensionality explosion, this time in the
number of grid cells. Additionally, such basic implementation truncates the
exploitation abilities of the global optimization algorithm. This causes the
algorithm with static niches to require more steps in those low dimensional
cases where the problem can also be solved without. From our experience with LJ
clusters, we expect more advanced niching strategies, for example dynamic grids,
to prove useful with higher dimensionalities and render the function even less
difficult. 

As it seems to be common wisdom in this area, applications do contain some
degree of deceptiveness and different degrees of minima correlation, as well
as different landscape characteristics, sampling all possibilities between
golf courses and funnels. All of that can be captured with the GRUNGE
setup. It thus offers all the necessary flexibility and simplicity, combined
in a single function definition. The obvious downsides are the absence of a
pre-defined global minimum (which can also be interpreted as a guarantee for
avoiding biases towards it) and the need to tune many parameters to achieve a
desired landscape shape.

\section{Summary and Outlook}

Scaling investigations for four different, standard benchmark functions have
been presented, supplemented by performance tests on a fifth function. Using
straightfoward GA techniques without problem-specific ingredients, the
behavior we observe in all these cases is markedly different from what we
observe upon applying the same techniques to real-world problems (often
including system-specific additions): All benchmarks can be solved with
sub-quadratic scaling, whereas in real-world applications we can get cubic
scaling at best and often have to settle for much worse. In addition, the
benchmarks often do not allow for statistically significant conclusions
regarding the performance of different crossover operators, nor for a decision
on whether to include local optimization or not. Thus, overall, benchmarks of
this type do not seem to fulfill their purpose of test beds with relevance for
practical applications in global cluster structure optimization or similar
areas. 

We have contrasted the behavior on those traditional benchmark functions with
that on two different types of functions. One type is the ``landscape
generator'' class, shown here in a particularly simple realization,
namely search landscapes generated by
randomly distributed Gaussians. Varying Gaussian characteristics (depth,
width, density, dimensionality), search space features can be tuned at will,
on the full scale between a ``mountain range'' and a ``golf course''. In
addition, since the constituting Gaussians are completely uncorrelated, the
difficulty of this problem class is inherently larger than that of the
traditional benchmark functions where the minima characteristics follow a
simple rule by construction. This is strikingly reflected in our tests results
on this benchmark class.

As yet another class of benchmark functions we have shown the deceptive type,
designed to lead global optimization astray. Their difficulty can be
diminished significantly by making the global search more sophisticated, in
the case of population-based searches by ensuring a sufficient degree of
diversity in the population. Given a sufficiently flexible setup, this
benchmark class merely is a subclass of the landscape generators.

Further work will be required to confirm our suspicion that real-world
problems often fall in-between the traditional, rather simple benchmark
functions on the one hand and the less correlated, more deceptive ones on the
other hand, both with respect
to their search space characteristics and to the difficulty they present for
global optimization algorithms. In any case, we hope to have shown
convincingly that due to their simplicity functions from the traditional
benchmark functions should not be used on their own, neither to aid global
optimization algorithm development nor to judge performance.

\end{document}